\documentclass[runningheads]{llncs}
\usepackage[T1]{fontenc}
\usepackage{graphicx}
\usepackage{booktabs}
\usepackage[misc]{ifsym}
\usepackage{hyperref}
\newcommand{\corr}{(\Letter)}
\usepackage{amsmath}

\usepackage{algorithm}
\usepackage{algpseudocode}
\algrenewcommand{\algorithmiccomment}[1]{\hskip3em\(\%\) \textit{#1}}
\usepackage{multirow}
\usepackage{amssymb}

\begin{document}

\title{Towards Continuous Power Forecasting: Practical Continual Learning for Real-World Energy Systems in Nonstationary Time Series}

\titlerunning{Practical Continual Learning for Real-World Energy Systems}

\author{Yujiang He\inst{1}\corr \and
Frederic Uhrweiller\inst{1} \and
Bernhard Sick\inst{1}}

\authorrunning{Y. He et al.}

\institute{Intelligent Embedded Systems, University of Kassel, 34121 Kassel, Germany \email{\{yujiang.he, frederic.uhrweiller, bsick\}@uni-kassel.de}}

\maketitle              

\begin{abstract}
Power forecasting models deployed in real-world energy markets must operate under nonstationary conditions, where data distributions continually evolve due to weather variability, infrastructure upgrades, and changing consumption behaviors.
In practice, these models face strict operational constraints: historical data may be limited for repeated retraining, and uninterrupted long-term service is often required.

This paper addresses these challenges by proposing the paradigm of Continuous Power Forecasting, which views power forecasting as a continual learning problem rather than a static offline task.
Based on an adaptive continual learning framework for regression, we systematically investigate the practical effectiveness of six representative continual learning approaches from three methodological categories. 
These approaches are evaluated under different realistic assumptions regarding data accessibility and update policies.
Experimental validation on real-world power datasets demonstrates that continual learning enables forecasting models to self-adapt to distributional drift, accumulate knowledge over time, and mitigate catastrophic forgetting without relying on large-scale historical data storage.
Beyond performance gains, our study provides practical insights into the stability and adaptation behaviors of different continual learning approaches under realistic operational constraints.
Overall, this work illustrates how continual learning can be pragmatically integrated into industrial power forecasting pipelines, offering a scalable and sustainable solution for long-term deployment in dynamic environments.

\end{abstract}

\section{Introduction}
\label{sec:intro}
Power forecasting is a core component of modern energy systems, supporting market operations and grid management. 
In real-world deployments, forecasting models must operate over long lifecycles under nonstationary conditions, where data distributions evolve due to changing weather patterns, infrastructure, and consumption behavior.
At the same time, operational constraints such as limited access to historical data and the requirement for uninterrupted service make repeated offline retraining impractical.

In practice, nonstationarity acts as both drifts in input distributions and changes in the underlying relationship between inputs and targets due to concept drift.
While collecting large training datasets can partially mitigate such effects, this approach is costly, slow, and often incompatible with data privacy protection requirements.
Conversely, naively fine-tuning models on incoming data leads to catastrophic forgetting, and this weakens long-term forecasting reliability.

We address these challenges through the paradigm of \textbf{Continuous Power Forecasting} (CPF), which treats power forecasting as a lifelong adaptive process with the support of continual learning (CL) rather than a static offline task.
Here, \textit{continuous} emphasizes uninterrupted system-level operation over the model’s lifecycle, while {continual} denotes the repetitive learning mechanism that enables incremental adaptation and knowledge retention.
The goal is to allow forecasting models to autonomously adapt to distributional drift and accumulate knowledge over time without relying on large-scale historical data storage.

Based on the modular CLeaR (\textbf{C}ontinual \textbf{Lea}rning for \textbf{R}egression) framework~\cite{he2021clear}, which enables unified comparison of distinct continual learning approaches for regression, this paper investigates practical effectiveness of six practical CL approaches for power forecasting under realistic operational assumptions.
The main contributions in this work can be summarized as follows:
\begin{itemize}
    \item We formalize CPF as an operational paradigm for lifelong energy forecasting systems operating under nonstationary conditions and strict constraints.

    \item We conduct a unified large-scale empirical evaluation of six representative continual learning approaches under realistic data accessibility constraints, providing mechanism-level insights into their stability-plasticity trade-offs across diverse power entities.
    
    \item We ensure reproducibility by evaluating all implementations on publicly accessible datasets and releasing the code of all evaluated CLeaR instances~\footnote{
    \url{https://github.com/HYJrrr/CLeaR\_Framework}}.
\end{itemize}

The remainder of this paper is organized as follows. 
Section \ref{sec:related_work} reviews related work on continual learning and adaptive forecasting. 
Section \ref{sec:clear_instance} introduces the CLeaR framework and various continual learning instantiations. 
Section \ref{sec:experiments} describes the experimental setup and evaluation protocol, followed by a comparative analysis of results.
Section \ref{sec:conclusion} concludes with design guidelines for CPF systems and directions for future work.

\section{Related Work}
\label{sec:related_work}
CL studies how models can learn multiple tasks in sequence without suffering catastrophic forgetting, a phenomenon fundamentally rooted in the stability-plasticity dilemma ~\cite{wang2024survey}. 
A large body of CL research categorizes solutions into three main groups~\cite{de2021continual}: regularization-based approaches~\cite{kirkpatrick2017overcoming,zenke2017continual,schwarz2018progress} that constrain parameter updates, replay-based approaches~\cite{rolnick2019experience,rebuffi2017icarl,shin2017continual,van2018generative} that utilize stored or synthesized historical samples to anchor optimization, and architecture-based approaches~\cite{rusu2016progressive} that assign dedicated parameter subspaces to different tasks to eliminate interference.
However, standardized CL benchmarks mainly focus on classification settings~\cite{de2021continual,hsu2018re,lomonaco2017core50}, leaving open questions about the applicability of these approaches to forecasting scenarios, which remain comparatively less explored in the CL literature~\cite{he2021clear,cossu2020continual}.

In the specific domain of power systems, continuous forecasting is a practical necessity due to evolving consumption patterns, generation variability, and operational constraints~\cite{he2020continuous}. 
While state-of-the-art forecasting architectures, such as Informer~\cite{zhou2021informer}, Autoformer~\cite{wu2021autoformer}, and PatchTST~\cite{nie2022time}, have achieved remarkable accuracy on static benchmarks, they mostly rely on offline training or periodic retraining on large-scale historical datasets.
Such assumptions are often infeasible in real deployments with restricted data accessibility, privacy constraints, or uninterrupted service requirements.

Related literature on adaptive time-series forecasting encompasses methods for concept drift detection and model updating, including statistical change detection and drift-aware learning~\cite{gama2014survey,lu2018learning}, online learning algorithms~\cite{sahoo2017online}, and ensemble-based adaptation techniques~\cite{kolter2007dynamic}. 
These approaches aim to maintain predictive performance under evolving distributions, but typically emphasize detection and reactive retraining rather than explicit stability-plasticity trade-offs under memory constraints.

Bounded-memory continual regression has recently been instantiated in modular frameworks such as CLeaR~\cite{he2021clear,he2021toward}, which integrates buffer-based storage, novelty detection, and multiple update strategies for streaming settings. 
However, systematic empirical evaluation of heterogeneous CL approaches under realistic operational constraints remains limited.

Our work builds on these strands by providing a unified empirical evaluation of multiple CL approaches within a consistent protocol on a real-world power grid dataset. 
By situating regularization-based, replay-based, and generative approaches in a common framework and analyzing their behavior under diverse nonstationary conditions, we offer insights that extend beyond static benchmark settings.
This perspective contributes to both the CL literature and the applied data science community by clarifying the conditions under which particular adaptation strategies may be preferred in dynamic, resource-constrained forecasting environments.
\section{CLeaR Framework and Instantiations}
\label{sec:clear_instance}
CLeaR is an adaptive and modular CL framework designed for regression problems, where forecasting models must evolve to adapt to changing data distributions over time under streaming data and limited historical storage constraints~\cite{he2021clear}.
At the framework level, CLeaR consists of two main modules: a buffer-based CL module and a novelty detection module.
The CL module mainly contains: (1) a neural network-based forecasting model, (2) familiarity and novelty buffers, (3) a CL approach used to update model parameters, and (4) a trigger mechanism.
The familiarity buffer stores samples consistent with the current model state, while the novelty buffer accumulates samples indicating distributional drift in the data stream.
These buffer types are intrinsic to the CLeaR framework, as continuous monitoring of potential task changes is required for controlled adaptation.
Additional buffers, such as replay buffers, are optional and depend on the specific CL approach adopted in a given instantiation (see Section~\ref{subsec:cl}).
The novelty detection module assigns incoming samples to buffers according to a detection function that quantifies distributional deviation relative to the deployed model.

\begin{figure}[t]
\includegraphics[width=\textwidth]{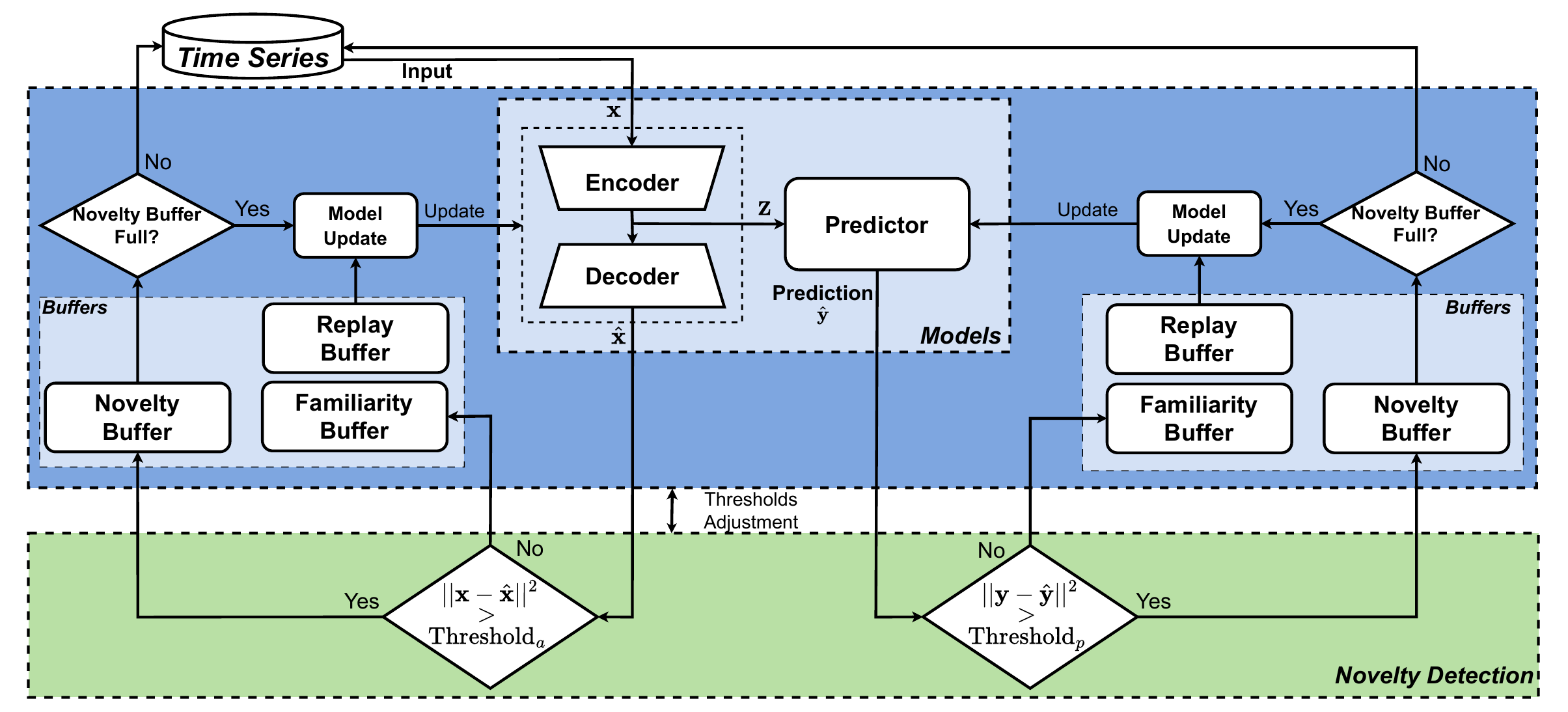}
\caption{CLeaR instance with real replay, showing dual-branch novelty detection (reconstruction and prediction errors), buffer allocation, and asynchronous model update.} 
\label{fig:clear_replay}
\end{figure}

Operationally, CLeaR follows a novelty-driven streaming adaptation loop. 
Given a data stream $\{\mathbf{x}_t, y_t\}_{t=1}^{T}$ under bounded memory constraints, CLeaR iteratively: (1) receives streaming samples; (2) generates forecasts and evaluates detection signals; (3) assigns samples to appropriate buffer(s); (4) triggers parameter updates when predefined conditions are satisfied; (5) replaces the deployed model with the updated version, and  (6) continues monitoring under the new model state.
Notably, these components are instantiated according to application requirements.
A concrete implementation of CLeaR is referred to as a CLeaR instance, obtained by specifying each defined module.
Figure~\ref{fig:clear_replay} illustrates an instance used in this study.
While the figure depicts a real-replay instantiation, the detection and update mechanism described below is shared across all instances unless otherwise specified.

The following subsections describe the instantiation of the adopted detection and update mechanism, and detail the evaluated CL approaches.

\subsection{Novelty Detection and Update Mechanism}
\label{subsec:novelty_detection}
All evaluated CLeaR instances share a common forecasting architecture composed of an autoencoder (AE) for representation learning, and a target-specific multilayer perceptron predictor.
The AE is first trained to learn compact representations of input features.
Once training is completed, the encoder parameters are frozen, and the learned representation is fed into the predictor for supervised training. 
This architecture is intentionally modular and extendable.
The neural network can be replaced by more advanced neural networks, extended for multi-task learning, or substituted with a variational autoencoder (VAE) when Generative Replay is required.

The model provides two measurable error signals: reconstruction and prediction errors. 
The former reflects drifts in the input distribution, while the latter captures deviations in the input-target relationship.
These signals form the basis of novelty detection in this instantiation.
At each time step, the deployed model generates forecasts and computes both errors. 
Each error is compared against a dynamically adjusted threshold. 
These thresholds are defined as a scaled running mean error of the most recently updated model.
Specifically, after each update, the threshold is re-estimated as $\tau = \alpha \cdot \bar{e}$, where $\bar{e}$ denotes the mean reconstruction or prediction error evaluated on the samples currently stored in the familiarity and novelty buffers, and $\alpha$ is a user-defined scaling factor controlling sensitivity.
Smaller $\alpha$ increases sensitivity to deviations, while larger values reduce unnecessary updates.
Thresholds are initialized using the full warm-up dataset during the warm-up phase and then adjusted only after each model update.
Errors on familiarity samples reflect stability, whereas errors on novelty samples reflect adaptation quality. 
The re-estimated threshold therefore implicitly reflects the current stability-plasticity balance achieved by the updated model.

Samples exceeding the reconstruction or prediction threshold indicate potential drifts in the input distribution and in the underlying input-target relationship, respectively. 
Input samples $\mathbf{x}$ are stored in the corresponding familiarity or novelty buffer of the autoencoder, while labeled pairs $\{\mathbf{x}, y\}$ are assigned to the buffers associated with the predictor. 
This separation enables independent monitoring of distributional and conditional shifts.
Each novelty buffer is associated with a predefined capacity.
When the number of stored novelty samples reaches this capacity, the corresponding model component is scheduled for update.
Through the capacity-based trigger, the updating frequency is jointly determined by the model’s performance, the sensitivity of the detection thresholds, and the magnitude of distributional drift in the data stream.
Once the model update is finished, the updated model replaces the previously deployed model.
Thereafter, the corresponding threshold is re-estimated and the corresponding buffers are then emptied.
Streaming monitoring continues under the new model state until the next update is triggered.
In this study, updating is performed asynchronously for most instantiations.
The autoencoder and predictor maintain independent buffers and are updated only when their respective novelty buffers reach capacity. 
It enables differentiated responses to distributional and conditional drifts.
For Generative Replay, however, synchronous updates are adopted to ensure consistency between representation learning and generated data.

Different CLeaR instances share this workflow, while differing in the CL approaches used during parameter updating, as described in Section~\ref{subsec:cl}.

\subsection{Continual Learning Approaches}
\label{subsec:cl}
Rather than reiterating the standard taxonomy of continual learning, we focus on how different update mechanisms are instantiated within CLeaR under operational CPF constraints.
Specifically, we study three update strategies that differ in how historical information is preserved and accessed during parameter adaptation: (1) regularization-based weight consolidation, (2) real replay from explicitly stored historical samples, and (3) pseudo replay via learned data approximation through generated or curated samples.
These approaches differ in memory footprint, computational overhead, and their assumptions regarding historical data accessibility, which are critical under streaming CPF settings.
For convenience, each approach is instantiated as a CLeaR instance for experimental validation (Section~\ref{sec:experiments}), with abbreviations for each instance introduced in Section~\ref{subsec:cl_protocol}.

\subsubsection{Online EWC}
Elastic Weight Consolidation (EWC)~\cite{kirkpatrick2017overcoming} mitigates catastrophic forgetting by constraining updates to parameters that are important for previously learned data. 
Instead of storing historical samples, EWC estimates parameter importance using the Fisher Information (FI) and penalizes deviations from previously consolidated parameter values.

In the online variant~\cite{schwarz2018progress}, parameter importance is accumulated recursively. 
Let $\theta^{*}_{i}$ denote the $i$-th consolidated weight parameter from the previous update and $F^{*}_{i}$ the corresponding accumulated FI. 
When the trigger condition is fulfilled, the model parameters are optimized by minimizing:
\begin{equation}
\mathcal{L}_{\text{EWC}}(\theta) 
= \mathcal{L}_{\text{N}}(\theta; \mathcal{D}_N) 
+ \frac{\lambda}{2} \sum_i F^{*}_i (\theta_i - \theta^{*}_i)^2,
\end{equation}
where $\mathcal{L}_{\text{N}}$ denotes the task loss solely on the novelty samples $\mathcal{D}_N$, which is interpreted as representing newly emerged task information, and $\lambda$ controls the strength of regularization.
After optimization, the accumulated FI is updated recursively: $F^{*} = \gamma F^{*} + F_{\text{new}}$, where $F_{\text{new}}$ is estimated using both the novelty and familiarity buffers since historical data are assumed no longer fully accessible.
This design implicitly assumes that the familiarity buffer provides a sufficient summary of recent historical distributions for consolidation purposes.
$\gamma \in [0,1]$ is a decay factor controlling the retention of historical importance.

\subsubsection{Real Replay-based Approaches}
In real replay, historical samples are explicitly stored and replayed during parameter updates.
Real replay-based approaches mitigate forgetting by replaying a subset of previously observed samples and jointly optimizing on replayed and novelty data.
When an update is triggered at time $t_{k}$, the replay buffer $\mathcal{D}_R$ is populated by sampling a fixed number of real historical samples from all data observed prior to the previous update time $t_{k-1}$.
Importantly, the replay buffer is reconstructed at each trigger event.
Let $\mathcal{D}_N$ denote the novelty dataset collected between $t_{k-1}$ and $t_{k}$.
Model parameters are optimized by minimizing the combined objective:
\begin{equation}
\mathcal{L}_{\text{replay}}(\theta)=\mathcal{L}_{\text{N}}(\theta; \mathcal{D}_N)+\lambda_R \mathcal{L}_{\text{R}}(\theta; \mathcal{D}_R),
\label{eq:replay_loss}
\end{equation}
where $\mathcal{L}_{\text{N}}$ and $\mathcal{L}_{\text{R}}$ denote the task loss evaluated on novelty and replay samples respectively, and $\lambda_R$ controls the relative contribution of replayed data.
A fixed replay size is adopted for computational tractability and bounded memory usage, making deployment in real-world streaming scenarios feasible.
The replay approaches differ only in how $\mathcal{D}_R$ is constructed under memory constraints.
Detailed algorithmic procedures and pseudocode are provided in the Appendix. 

\paragraph{Random Replay}
This constructs $\mathcal{D}_R$ at each update trigger by uniformly selecting samples from the historical data available prior to the last completed update. 
Each eligible historical sample has equal probability of being included in the replay set, irrespective of its recency.
Under a fixed replay size, Random Replay can be interpreted as a temporal downsampling mechanism.
Given that weather and power time-series evolve gradually, such downsampling preserves the overall temporal trends while reducing short-term fluctuations.
It serves as a baseline mechanism without imposing temporal preference.

\paragraph{Recent Replay}
To construct $\mathcal{D}_R$, this approach uniformly selects samples from the historical data observed during the $N_R$ most recent updates.
Formally, if $n$ updates have been completed, $\mathcal{D}_R$ comprises samples from updates $n-N_R+1$ to $n$, or all available updates if $n < N_R$.
By prioritizing recent updates, this approach emphasizes short-term stability while allowing older knowledge to gradually fade. 
When the given $N_R$ exceeds the total number of currently completed updates, Recent Replay reduces to Random Replay over the historical data, providing an empirical unbiased approximation of the consolidated past experiences.

\paragraph{Recent Replay with Decay}
This approach extends the standard Recent Replay by introducing a recency-based weighting over the most recent $N_R$ updates.
Within this temporal window, samples associated with more recent updates are assigned higher selection probabilities, linearly increasing with recency.
During replay set construction, the number of samples drawn from each update is proportional to its weight, ensuring that recent knowledge is emphasized while still retaining a limited influence from older updates.
If the total number of updates $n$ is smaller than $N_R$, the weighting is applied over all available updates, yielding a gradually decayed sampling distribution across the historical dataset.

\subsubsection{Pseudo Replay-based Approaches}
Instead of maintaining a real replay buffer, pseudo replay generates or curates samples representing past knowledge, which are then incorporated into parameter updates.
It reduces direct memory dependence while preserving past information through implicit distribution modeling.

\paragraph{Generative Replay}
This implements pseudo replay by replacing the AE with a VAE to synthesize historical samples during adaptation.
Unlike a deterministic AE, the VAE encoder regularizes the latent space toward a standard prior, enabling the decoder to synthesize diverse pseudo-samples by sampling from this learned distribution.

After each completed update, the model parameters are archived to preserve the consolidated historical state.
When a new update is triggered, pseudo historical sample pairs are constructed using these archived models.
Specifically, latent codes are sampled from the approximated prior distribution.
The archived decoder generates pseudo input sequences from these latent representations, while the archived predictor simultaneously produces the corresponding pseudo outputs from the same representations.
As a result, replay samples appear as paired input-output, ensuring consistency between reconstructed features and targets.

For updating, the generated pseudo dataset $\mathcal{D}_G$ is combined with the novelty dataset $\mathcal{D}_N$.
Model parameters are optimized using the objective that mirrors Eq.~\ref{eq:replay_loss} but replaces real replay data with generated pseudo dataset.
Because pseudo input and output are generated jointly from shared latent variables, both the VAE and the predictor must be updated synchronously.
This synchronized optimization preserves alignment between representation learning and predictive mapping, which is critical for maintaining coherent historical knowledge across adaptation steps.
Compared to real replay, generative replay reduces explicit storage requirements at the cost of additional model training and sampling overhead, offering a flexible memory-compute trade-off under constrained storage scenarios.

\paragraph{Familiarity-based Replay}
This approach utilizes familiarity samples to strengthen parameter updates using EWC.
Unlike other CLeaR instantiations, where familiarity samples are only used for FI estimation and threshold adjustment, this approach incorporates $\widetilde{\mathcal{D}}_F$, a subset from the familiarity buffer, into the model optimization.
When an update is triggered, the optimization objective extends Eq.~(1) by replacing $\mathcal{D}_N$ with combined updating dataset $\mathcal{D}_{\text{comb}} = \mathcal{D}_N \cup \widetilde{\mathcal{D}}_F$.
The proportion of familiarity samples is controlled to prevent excessive dominance over novelty-driven adaptation.
The Fisher coefficients $F^{*}_i$ remain estimated from previous updates and are not modified by familiarity data.
The inclusion of familiarity samples introduces an additional optimization anchor and complement the parameter-space regularization imposed by EWC.
Since both buffers participate in the update, this approach also mitigates potential misallocation errors arising from imperfect novelty detection by retaining samples deemed stable under the current model.
\section{Experimental Validation of Continuous Power Forecasting}
\label{sec:experiments}
\subsection{Real-World Power Grid Dataset}
The experiments are conducted on a real-world regional power grid dataset (available upon reasonable request).
The dataset spans approximately 23 months with a uniform temporal resolution of 15 minutes, resulting in 67\,764 time steps.
The dataset comprises 95 power entities, including 7 independent generators, 59 independent consumers, and 29 aggregated residual loads. 
The residual loads represent consumption units without individual measurements and are estimated from substation-level aggregated statistics. 
All entities are anonymized due to privacy protection.

Forecasting is formulated as a pointwise mapping conditioned on given meteorological inputs rather than as a sequence-to-sequence task.
The meteorological data are derived from a day-ahead numerical weather prediction (NWP) system and include 13 variables such as wind speed, temperature, radiation, cloud cover, and precipitation. 
Given the relatively small geographical area (approximately 83~km$^2$), spatially uniform weather inputs are assumed for all entities, and spatial variance is not explicitly modeled.
The original hourly NWP data are linearly interpolated to match the 15-minute resolution of the power measurements.
The interpolation is applied exclusively to NWP and does not involve future observed values.
In addition, five timestamp-based temporal indicators (hour, minute, week, weekday, day-of-year) are encoded using sine-cosine transformations, resulting in 10 cyclical temporal features. 
The final input dimensionality is therefore 23.

Power measurements are normalized by rated or peak capacity for each entity, and weather variables are scaled to the range $[0,1]$ based on historical extrema.

\subsection{Continual Learning Protocol}
\label{subsec:cl_protocol}
For clarity in the experimental results, each continual learning approach is instantiated as a corresponding CLeaR instance and abbreviated as follows: Online EWC is referred to as $\text{CLeaR}_{\text{E}}$, Random Replay as $\text{CLeaR}_{\text{Ra}}$, Recent Replay as $\text{CLeaR}_{\text{Re}}$, Recent Replay with Decay as $\text{CLeaR}_{\text{ReD}}$, Generative Replay as $\text{CLeaR}_{\text{G}}$, and Familiarity-based Replay as $\text{CLeaR}_{\text{F}}$. 
The data for each power entity is divided into three phases: warm-up, updating, and test. 
\begin{table}[!b]
    \centering
    \caption{Grid search space for AE/VAE and predictor. The encoder begins with the specified number of neurons in the first hidden layer, with subsequent layers containing 70\% of the neurons in the previous layer. The decoder mirrors the encoder to reconstruct input features.}
    \begin{tabular}{c|c|c}
        \toprule
        \textbf{Model} & \textbf{Hyperparameter} & \textbf{Candidate Values}\\
        \midrule
        \multirow{3}{*}{\centering AE and VAE} 
        & Number of Hidden Layers & \{3, 4\}\\
        \cline{2-3}
        & Neurons in the 1st Hidden Layer & \{64, 128, 256\}\\
        \cline{2-3}
        & Latent Dimension $d_{\mathbf{z}}$ & \{18, 16, 14, 12\}\\
        \hline
        VAE & KL Divergence Weight & \{$5\times 10^{-5}$, $5\times 10^{-6}$,$5\times 10^{-7}$\}\\
        \hline
        \multirow{3}{*}{\centering \shortstack{Predictor}}& Input Size & Latent Dimension $d_{\mathbf{z}}$\\
        \cline{2-3}
        & Number of Hidden Layers & \{3, 4\}\\
        \cline{2-3}
        & Neurons per Hidden Layer & \{64, 128, 256\}\\
        \bottomrule
    \end{tabular}
    \label{tab:grid_search_model}
\end{table}

During the warm-up phase, the first 10\,000 samples (approximately 104 days) are used to initialize the shared AE (or VAE for $\text{CLeaR}_{\text{G}}$) and entity-specific predictors, identifying optimal architectures and latent dimensions.
This phase allows the model to capture fundamental patterns from limited historical data.
All tunable parameters for AE/VAE and predictor are specified in Table~\ref{tab:grid_search_model}. 
Warm-up training uses 80\% of warm-up samples for training and 20\% for validation, with early stopping (patience 50), a maximum of 512 epochs, batch size 64, and the Adam optimizer with learning rate 0.001.
Hyperparameter selection is based on the lowest validation root mean squared error (RMSE), using reconstruction error for the AE/VAE and prediction error for the entity-specific predictor.
\begin{table}[!t]
    \centering
    \caption{Grid search space for hyperparameters of CLeaR instances during the updating phase. Replay buffer size for all replay-based instances (Ra, Re, ReD) equals the novelty buffer size.}
    \begin{tabular}{c|c|c}
        \toprule
        \textbf{Instance} & \textbf{Hyperparameter} & \textbf{Candidate Values}\\
        \midrule
        \multirow{2}{*}{\centering All Instances} & Novelty Buffer Size & \{3\,000, 5\,000\}\\
        \cline{2-3}
        &Threshold Factor $\alpha$& \{0.9, 1.3\}\\
        \hline
        OEWC / Familiarity & Regularization Coefficient $\lambda_{\mathrm{EWC}}$& \{$10^{4}$, $5\times 10^{4}$, $5\times 10^{5}$\}\\
        \hline
        Ra / Re / ReD/ Generative & Replay Coefficient $\lambda_{\mathrm{R}}$& \{0.5, 1.0, 1.5\}\\
         \hline
        Re / ReD & Number of Tasks to Replay & \{3, 5, 7\}\\
        \bottomrule
    \end{tabular}
    \label{tab:grid_search_clear_updating}
\end{table}

Following the warm-up, each CLeaR instance enters the updating phase, spanning 54\,764 samples over approximately 570 days. 
During this phase, instances continuously generate forecasts while monitoring detection signals from incoming samples. 
When a novelty buffer reaches capacity, the corresponding neural network component is updated according to the selected CL approach. 
All hyperparameters controlling these updates are specified via grid search and summarized in Table~\ref{tab:grid_search_clear_updating}.
For each CLeaR instance, hyperparameter configurations in the updating phase are selected individually for each entity based on the lowest RMSE on the updating dataset (later referred to as Fitting Error).
No additional regularization is applied during this phase, as the CL approaches inherently introduce functional regularization via EWC penalties or replay losses.

The updating phase is followed by the test phase, where the remaining 3\,000 samples (approximately 31 days) are used to evaluate the final forecasting performance of each CLeaR instance.
\textbf{Two baselines} are included: $\text{Baseline}_\text{L}$, representing the warm-up model without further updates, and $\text{Baseline}_\text{U}$, trained on the full dataset spanning both warm-up and updating phases, serving as empirical lower and upper performance references, respectively. 
This protocol ensures fair, reproducible, and realistic evaluation under operational constraints typical of industrial power forecasting pipelines.

\subsection{Evaluation Metrics}
\label{subsec:metrics}
Each CLeaR instance is evaluated after the updating phase using three post-update metrics~\cite{he2021clear}: Fitting Error (FE), Prediction Error (PE), and Forgetting Ratio (FR).
All metrics are computed on the fixed warm-up, updating, and test datasets using the final model obtained after the last update.
Let $\mathrm{M}_{\mathrm{final}}$ denote the final model obtained at time $t_{\mathrm{final}}$, and $\mathrm{M}_{\mathrm{final}}\!\left[\mathbf{x}(t)\right]$ its output for input $\mathbf{x}(t)$. 
We introduce a unified target notation $\mathbf{d}(t)$, where $\mathbf{d}(t)=\mathbf{x}(t)$ for reconstruction and $\mathbf{d}(t)=y(t)$ for forecasting.

FE measures how well the final model fits all samples that have been seen during the updating phase. 
It is defined as the RMSE over the corresponding temporal interval:
\begin{align}
     \mathrm{FE} = \sqrt{\frac{\sum_{t=10\,001}^{t_{\mathrm{final}}}\left(\mathbf{d}\left(t\right)- \mathrm{M}_{\mathrm{final}}\left[\mathbf{x}\left(t\right)\right]\right)^2}{t_{\text{final}}-10\,000}}.
     \label{eq:fe}
\end{align}
Samples collected after $t_{\mathrm{final}}$ but not triggering an additional update are excluded from this calculation, as they do not influence model parameters.

PE evaluates the generalization performance of the final model on unseen data from the test phase:
\begin{align}
    \mathrm{PE} &= \sqrt{\frac{\sum_{t=64\,765}^{67\,764}\left(\mathbf{d}\left(t\right)- \mathrm{M}_{\mathrm{final}}\left[\mathbf{x}\left(t\right)\right]\right)^2}{3\,000}}.
    \label{eq:pe}
\end{align}
PE reflects out-of-sample performance beyond both warm-up and updating data. 

FR quantifies the relative increase in error on the warm-up dataset after the completion of the updating phase. 
Let $\mathrm{M}_{0}$ denote the warm-up model (i.e., $\text{Baseline}_\text{L}$).
FR is defined as:
\begin{align}
    \mathrm{FR} &=\mathrm{max}\left\{0,\  \sqrt{\frac{\sum_{t=1}^{10\,000}\left(\mathbf{d}\left(t\right)- \mathrm{M}_{\mathrm{final}}\left[\mathbf{x}\left(t\right)\right]\right)^2}{\sum_{t=1}^{10\,000}\left(\mathbf{d}\left(t\right)- \mathrm{M}_{\mathrm{0}}\left[\mathbf{x}\left(t\right)\right]\right)^2}} -  1 \right\}.
    \label{eq:fr}
\end{align} 
If the final model performs no worse than the warm-up model on the warm-up dataset, FR equals zero. 

Together, FE, PE, and FR provide complementary perspectives on continual regression performance, respectively capturing fitting quality on updated data, generalization to unseen data, and resistance to catastrophic forgetting.

\subsection{Overall Performance Comparison}
Using the optimal configurations selected as described in Section~\ref{subsec:cl_protocol}, we report aggregated performance statistics across all 95 entities.

\paragraph{AE Component (Input Feature Reconstruction)}
Table~\ref{tab:results_overall_ae} summarizes the post-update evaluation results for the AE component across all CLeaR instances. 
Since the baseline models are not updated during the updating phase, the values of FR and number of updates are not applicable in the corresponding rows.

As expected, $\text{Baseline}_{\text{L}}$, which is trained only on the limited warm-up dataset, exhibits substantially higher FE and PE during later phases. 
All CLeaR instances achieve markedly lower FE than $\text{Baseline}_{\text{L}}$, indicating that continual updates substantially improve retention of input feature structure.

Clear differences exist between regularization-based and replay-based approaches. 
The four replay-based variants ($\text{CLeaR}_{\text{Ra}}$, $\text{CLeaR}_{\text{Re}}$, $\text{CLeaR}_{\text{ReD}}$, and $\text{CLeaR}_{\text{G}}$) consistently outperform $\text{CLeaR}_{\text{E}}$ and $\text{CLeaR}_{\text{F}}$ in terms of all three metrics. 
This suggests that explicit replay of historical samples outperforms regularization-based constraints in preserving both reconstruction accuracy and the stability of feature-level representations.
\begin{table}[!t]
    \centering
    \caption{Post-update evaluation of AE component (input feature reconstruction). Results are aggregated across 95 power entities and presented as mean and standard deviation (in parentheses). For reference, the mean errors achieved by AE on validation data for $\text{Baseline}_{\text{L}}$ and $\text{Baseline}_{\text{U}}$ are $0.005$ and $0.003$, respectively. $\text{Baseline}_{\text{L}}$ and $\text{Baseline}_{\text{U}}$ values in the table represent their performance during the updating/test phases.}
    \resizebox{\textwidth}{!}{
        \begin{tabular}{c|c|c|c|c}
            \toprule
            \textbf{Instances} & \textbf{Fitting Error} & \textbf{Prediction Error} & \textbf{Forgetting Ratio} & \textbf{Number of Updates}\\
            \midrule
            $\text{Baseline}_{\text{U}}$ & 0.003 & 0.021  & / & / \\
            \hline
            $\text{Baseline}_{\text{L}}$ & 0.345 &  0.203 & / & / \\
            \hline
            $\text{CLeaR}_{\text{E}}$ & 0.249 (0.037) & 0.180 (0.072)& 91.270 (13.061) & 13.368 (3.516)\\
            \hline
            $\text{CLeaR}_{\text{F}}$ & 0.243 (0.037)& 0.208 (0.094)& 90.365 (13.612) & 12.358 (3.418)\\
            \hline
            $\text{CLeaR}_{\text{Ra}}$ & 0.014 (0.001) & 0.053 (0.004) & 0.669 (0.123) & 8.442 (1.194)\\
            \hline
            $\text{CLeaR}_{\text{Re}}$ & 0.016 (0.008)& 0.057 (0.019) & 1.503 (1.457) & \textbf{8.253} (0.725)\\
            \hline
            $\text{CLeaR}_{\text{ReD}}$ & 0.016 (0.010)& 0.058 (0.023) & 1.533 (1.418) & 8.895 (0.788)\\
            \hline
            $\text{CLeaR}_{\text{G}}$ & \textbf{0.011} (0.001) & \textbf{0.045} (0.005) & \textbf{0.551} (0.215) & 13.484 (3.008)\\
            \bottomrule
        \end{tabular}
        }
    \label{tab:results_overall_ae}
\end{table}

Two factors may partly explain the inferior performance of $\text{CLeaR}_{\text{E}}$ and $\text{CLeaR}_{\text{F}}$. 
First, hyperparameter selection prioritizes forecasting FE rather than reconstruction performance.
Therefore, configurations optimal for predictive accuracy do not necessarily preserve the AE’s latent representation. 
Second, without access to historical samples, $\text{CLeaR}_{\text{E}}$ and $\text{CLeaR}_{\text{F}}$ must rely mainly on FI-based regularization to balance stability and plasticity. 
Under limited model capacity, which is intentionally constrained during warm-up to prevent overfitting, frequent updates (approximately 13 on average) increase the risk of overwriting the overlapped latent representations. 
In Online EWC, the down-weighting parameter $\gamma$ progressively relaxes older constraints, further leading to the relatively high FR.

While FR appears disproportionately large for regularization-based methods, this effect is amplified by normalization in Eq.~\ref{eq:fr}. 
Because the warm-up reconstruction error is very small (approximately 0.005 on average), even moderate absolute degradation results in a large relative ratio. 
Thus, FR magnifies representational drift when the baseline denominator is small.

Among all variants, $\text{CLeaR}_{\text{G}}$ achieves the best overall reconstruction performance, approaching the errors of $\text{Baseline}_{\text{U}}$, which is trained on the full dataset. 
This indicates that Generative Replay closely approximates full-data retraining under streaming constraints. 
Notably, $\text{CLeaR}_{\text{Ra}}$ also performs competitively despite employing a simple random replay strategy, highlighting that even non-selective replay provides a strong anchor against representational drift.

Although $\text{CLeaR}_{\text{G}}$ exhibits a relatively high number of updates due to its synchronous triggering mechanism, its low FR and strong FE/PE indicate that Generative Replay can maintain stability even under frequent parameter adjustments.

\paragraph{Predictor Component (Power Forecasting)}
Table~\ref{tab:results_all_pred} summarizes the aggregated post-update evaluation metrics for all 95 power entities.
\begin{table}[!b]
    \centering
    \caption{Post-update evaluation of the predictor component (power forecasting) for all 95 power entities, comparing various CLeaR instances and baseline models. Results are presented as the mean and standard deviation (in parentheses). For reference, the mean errors achieved by predictor on validation data for $\text{Baseline}_{\text{L}}$ and $\text{Baseline}_{\text{U}}$ are 0.037 (0.020) and 0.041 (0.020), respectively. $\text{Baseline}_{\text{L}}$ and $\text{Baseline}_{\text{U}}$ values in the table represent their performance during the updating/test phases.}
    \resizebox{\textwidth}{!}{
        \begin{tabular}{c|c|c|c|c}
            \toprule
            \textbf{Instances} & \textbf{Fitting Error} & \textbf{Prediction Error} & \textbf{Forgetting Ratio} & \textbf{Number of Updates}\\
            \midrule
            $\text{Baseline}_{\text{U}}$ & 0.037 (0.017) & 0.142 (0.083)  & / & /\\
            \hline
            $\text{Baseline}_{\text{L}}$ & 0.157 (0.074) &  0.165 (0.102) & / & / \\
            \hline
            $\text{CLeaR}_{\text{E}}$ & 0.116 (0.059) & 0.140 (0.079)& 2.057 (1.427) & 6.147 (1.941)\\
            \hline
            $\text{CLeaR}_{\text{F}}$ & 0.112 (0.060)& \textbf{0.135} (0.076) & 2.081 (1.506) & \textbf{5.505} (2.257)\\
            \hline
            $\text{CLeaR}_{\text{Ra}}$ & 0.086 (0.043) & 0.159 (0.083) & 1.079 (1.099) & 7.011 (2.075)\\
            \hline
            $\text{CLeaR}_{\text{Re}}$ & 0.080 (0.040)& 0.161 (0.085) & 1.925 (1.537) & 6.874 (1.842)\\
            \hline
            $\text{CLeaR}_{\text{ReD}}$ & 0.083 (0.040)& 0.158 (0.082) & 2.116 (1.616) & 6.895 (1.861)\\
            \hline
            $\text{CLeaR}_{\text{G}}$ & \textbf{0.046} (0.025) & 0.152 (0.086) & \textbf{0.344} (0.587) & 13.484 (3.008)\\
            \bottomrule
        \end{tabular}
    }
    \label{tab:results_all_pred}
\end{table}

Consistent with our previous findings regarding input feature reconstruction, all CLeaR instances achieve lower FE than $\text{Baseline}_{\text{L}}$, indicating the necessity and effectiveness of continual updates once the data distribution departs from the initial warm-up temporal window.
Among all instances, $\text{CLeaR}_{\text{G}}$ yields the lowest FE and FR, indicating the strongest retention of previously acquired knowledge under continual updates.

However, a different trend emerges in the PE metrics: $\text{CLeaR}_{\text{E}}$ and $\text{CLeaR}_{\text{F}}$ consistently surpass the replay-based instances, achieving lower mean PE than $\text{Baseline}_{\text{U}}$.
This difference reflects the task-dependent nature of the stability-plasticity trade-off.
In the input feature reconstruction context, the AE deals with NWP data, which exhibits high-frequency but short-term fluctuations with periodic structure.
As shown in Table~\ref{tab:results_overall_ae}, $\text{Baseline}_{\text{U}}$ maintains comparable errors on the validation data ($\approx 0.003$) and in the updating phase ($\approx 0.003$), with a minimal PE (0.021). 
This suggests relatively stable reconstruction statistics where preserving historical knowledge (stability) is more important.
Consequently, in this context, these replay-based instances, which can directly replay real/pseudo historical samples, substantially outperform the EWC-based variants across all evaluation metrics by effectively anchoring the model to its initial reconstruction proficiency.

Conversely, the power forecasting context may provide gradual drift in the input-output mapping relationship.
For $\text{Baseline}_{\text{U}}$, the error rises from 0.041 during warm-up to 0.142 in the test phase, indicating considerable concept drift in power data.
In this context, the controlled forgetting facilitated by the Online EWC down-weighting parameter $\gamma$ may provide additional plasticity.
By decaying outdated constraints, these instances exhibit higher plasticity, allowing them to adapt more quickly to new distribution characteristics.
As a result, EWC-based instances achieve lower PE despite higher FR, suggesting that stronger retention is insufficient to guarantee superior generalization under distributional drift.

A similar context-dependent behavior is observed in the update frequency.
Compared with the AE component, the predictor exhibits fewer updates on average.
This difference arises because AE updates (in Table~\ref{tab:results_overall_ae}) are primarily driven by short-term input distribution changes, whereas predictor updates (in Table~\ref{tab:results_all_pred}) are triggered by drifts in the underlying input-output relationship.
Such mapping drift occurs less frequently but could require more substantial parameter adaptation when detected.
\section{Conclusion \& Future Work}
\label{sec:conclusion}
This study investigates the CLeaR framework for CPF under nonstationary data streams with restricted data accessibility.
Extensive experiments on a real-world power grid dataset demonstrate that the framework effectively mitigates the limitations imposed by constrained historical storage and evolving operating conditions.
Across 95 heterogeneous entities, the CLeaR instances consistently improve forecasting robustness compared with static baselines, suggesting the practical importance of structured continual adaptation in deployment scenarios.

Although Generative Replay achieves superior performance across most metrics, simple performance ranking provides only a partial view of continual learning behavior.
The experimental results reveal that performance appears to depend on the nature and intensity of nonstationarity.
In relatively stable regimes, mechanisms emphasizing memory retention tend to maintain stronger generalization.
In contrast, under conditional drift, controlled relaxation of historical constraints may facilitate faster adaptation.
These findings indicate that stronger knowledge retention does not necessarily imply better generalization, particularly when the underlying input-output relationship evolves over time.

From a system design perspective, two principles emerge.
First, the selection of a continual learning mechanism must account for site-specific computational and storage constraints.
Methods that cannot be executed within operational constraints offer limited practical value regardless of theoretical performance.
Second, mechanism selection should be aligned with the form of nonstationarity. 
For example, seasonal or cyclic variations could favor stability-oriented approaches with stronger memory preservation.
Gradual physical degradation could require steady adaptation with moderate forgetting. 
Operational drifts or newly introduced forecasting targets prioritize rapid plasticity and efficient re-initialization.
Continual learning systems for power applications should therefore be drift-aware and constraint-aware rather than universally stability-oriented.

Several directions are proposed for further investigation.
First, exploring the integration of pre-trained models and zero-shot learning may accelerate cold-start adaptation and reduce dependency on extended warm-up phases.
Second, incorporating human-in-the-loop feedback into novelty detection and update triggering mechanisms may help bridge the gap between purely data-driven adaptation and expert-informed operational knowledge.
Third, selective forgetting mechanisms warrant systematic investigation in the context of long-horizon power forecasting.
While the present work focuses primarily on preserving past knowledge, long-horizon power system operation suggests that not all historical patterns remain informative and relevant.
Intentional removal of obsolete or no longer relevant patterns may improve both memory efficiency and long-term generalization.

Advancing along these directions will move CPF research beyond static mechanism comparisons toward adaptive, context-aware continual learning systems capable of operating reliably in dynamic energy infrastructures.

\begin{credits}
\subsubsection{\ackname} 
This work was supported within the KonSEnz (03EI4087B) project, funded by BMWE: Deutsches Bundesministerium für Wirtschaft und Energie/German Federal Ministry for Economic Affairs and Energy.

\subsubsection{\discintname}
The authors have no competing interests to declare that are relevant to the content of this article.

\subsubsection{Statement on the Use of AI Tools.} Large Language Models (LLMs) were used solely to improve the readability and correct grammatical errors in this manuscript. All scientific content, such as research ideas, the conceptual framework, experimental design, analysis, and conclusions, was developed entirely by the authors. The authors take full responsibility for the integrity and originality of the work.
\end{credits}
%
%
%
\bibliographystyle{splncs04}
\bibliography{mybibliography}

\newpage
\section*{Appendix}
This appendix provides algorithmic details for the three real replay-based continual learning (CL) approaches introduced in Section~\ref{subsec:cl} of the main paper titled \textit{Towards Continuous Power Forecasting: Practical Continual Learning for Real-World Energy Systems in Nonstationary Time Series}.
While the main text focuses on the unified replay loss formulation, this appendix specifies the data structures and sampling procedures used to construct the replay dataset for each approach.

We first define the historical dataset and replay buffer used by all three real replay approaches.
The historical dataset $\mathcal{D}_H$ represents the complete collection of observed samples accumulated up to the current model update.
Assume the $n$-th model update is completed at time $t_n$. 
The historical dataset $\mathcal{D}_H$  is defined as:
$$\mathcal{D}_H = \left\{\left(\mathbf{x}\left(t\right), \mathbf{y}\left(t\right)\right) \mid t_{-1} \leq t \leq t_{n}, t,n\in \mathbb{N}\right\},$$
where $t$ denotes the time step when the corresponding input-output sample pair is obtained and $t_{-1}\equiv 0$ denotes the starting time.

The replay buffer stores a subset of historical samples used for experience replay during model updates.
The replay dataset is denoted by $\mathcal{D}_R \subseteq \mathcal{D}_H$.
The replay budget $S_R$ specifies the number of historical samples selected for replay at each update step, i.e., $|\mathcal{D}_R|=S_R$.
In all real replay approaches considered in this work, the replay budget $S_R$ is treated as a fixed constant.

\subsection*{Random Replay}
Random Replay represents the simplest approach. 
It assumes that all historical samples stored in $\mathcal{D}_H$ are equally important for retaining past knowledge.
Following the completion of a model update (at time $t_{n}$), Random Replay generates the replay dataset $\mathcal{D}_R$ by uniformly sampling $S_R$ data pairs from the entire historical dataset $\mathcal{D}_H$ without replacement.
This sampled replay dataset $\mathcal{D}_R$ is a down-sampled subset of the complete historical data.
Since $S_R$ is constant, the probability of selecting a sample corresponding to a very old task gradually decreases as $\mathcal{D}_H$ grows, indirectly reducing the reliance on accessing the most outdated data.

Algorithm~\ref{alg:random_replay} describes the procedure for generating $\mathcal{D}_R$.

\begin{algorithm}[htb]
\caption{Random Replay Set Generation}
\label{alg:random_replay}
\begin{algorithmic}[1]
\State \textbf{Input:} Historical dataset $\mathcal{D}_H = \{(\mathbf{x}(t), \mathbf{y}(t)) \mid t_{-1} < t \le t_{n}\}$, Replay buffer size $S_R$
\State \textbf{Output:} Replay dataset $\mathcal{D}_R$
\Procedure{GenerateRandomReplaySet}{$\mathcal{D}_H$, $S_R$}
\Comment{Executed after the $n$-th model training/update is completed. Let $t_{-1} \equiv 0$.}
\If{$|\mathcal{D}_H| \le S_R$}
\State $\mathcal{D}_R \gets \mathcal{D}_H$ \Comment{If history is smaller than buffer size, use all samples}
\Else
\State $\mathcal{D}_R \gets \text{UniformRandomSample}(\mathcal{D}_H, S_R)$ \Comment{Randomly sample $S_R$ data pairs from $\mathcal{D}_H$ without replacement}
\EndIf
\State \textbf{return} $\mathcal{D}_R$
\EndProcedure
\end{algorithmic}
\end{algorithm}
\subsection*{Recent Replay}
Recent Replay is a variant, which recognizes that in nonstationary environments, more recent data may be more representative of the current data distribution or subsequent tasks.
It prioritizes data from the most recent tasks.
This approach requires an additional parameter, $N_R$, which defines the number of the most recent updates whose associated data is considered for replaying. 
Specifically, this corresponds to the data collected within the time intervals delineated by the last $N_R$ model update points.
Only samples within this time window are eligible for selection.
Following a model update, $\mathcal{D}_R$ is generated by uniformly sampling $S_R$ data pairs from the recent subset for use at the next model update.

\begin{algorithm}[htb]
\caption{Recent Replay Set Generation}
\label{alg:recent_replay}
\begin{algorithmic}[1]
\State \textbf{Input:} Historical dataset $\mathcal{D}_H = \{(\mathbf{x}(t), \mathbf{y}(t)) \mid t_{-1} < t \le t_{n}\}$, Replay buffer size $S_R$, Number of recent updates $N_R$, Number of updates $n$
\State \textbf{Output:} Replay dataset $\mathcal{D}_R$
\Procedure{GenerateRecentReplaySet}{$\mathcal{D}_H$, $S_R$, $N_R$, $n$}
\Comment{Executed after the $n$-th model training/update is completed. Let $t_{-1} \equiv 0$.}
\If{$n \leq N_R-1$}
\State $\mathcal{D}_{\text{Recent}} \gets \{(\mathbf{x}(t), \mathbf{y}(t)) \in \mathcal{D}_H \mid t_{-1} \le t \le t_{n}\}$ \Comment{Sampling starts from the initial time step $t=t_{-1}$}
\Else
\State $\mathcal{D}_{\text{Recent}} \gets \{(\mathbf{x}(t), \mathbf{y}(t)) \in \mathcal{D}_H \mid t_{n-N_R} \le t \le t_{n}\}$ \Comment{Samples associated with the last $N_R$ update intervals}
\EndIf
\State $N_{\text{Recent}} \gets |\mathcal{D}_{\text{Recent}}|$
\If{$N_{\text{Recent}} \le S_R$}
\State $\mathcal{D}_R \gets \mathcal{D}_{\text{Recent}}$ \Comment{If the recent subset is small, use all samples}
\Else
\State $\mathcal{D}_R \gets \text{UniformRandomSample}(\mathcal{D}_{\text{Recent}}, S_R)$ \Comment{Randomly sample $S_R$ data pairs from $\mathcal{D}_{\text{Recent}}$}
\EndIf
\State \textbf{return} $\mathcal{D}_R$
\EndProcedure
\end{algorithmic}
\end{algorithm}
When the number of conducted updates $n$ is less than $N_R$, the workflow is identical to Random Replay.
However, starting from $n=N_R+1$, the mechanism enforces a sliding window. 
Data corresponding to the oldest update outside the $N_R$ windows are sequentially excluded and thus will not be involved in subsequent replay generations.
By focusing on recent experiences, Recent Replay allows the model to adapt more quickly to recent changes in the underlying data distribution.
It effectively introduces a controlled mechanism to remove memories related to excessively outdated tasks.

Algorithm~\ref{alg:recent_replay} describes the Recent Replay sampling process. 

\subsection*{Recent Replay with Decay}
Recent Replay with Decay is an extension of Recent Replay that refines the sampling strategy by utilizing a simple yet effective mechanism to proportionally prioritize the influence of more recent samples while decaying the impact of older ones.
While this approach still focuses on the data associated with the most recent $N_R$ updates, it samples the historical data in a way that samples closer to the current update receive a proportionally higher sampling weight, thereby securing a greater share of the total replay budget $S_R$.

The approach assigns a weight to the data associated with each of the $N_R$ recent updates, typically decaying linearly. 
For instance, the data corresponding to the most recent update receives a weight of $N_R$, while the data for the oldest update within the window receives a weight of $1$.
The total replay budget $S_R$ is then proportionally distributed across these $N_R$ update intervals based on their assigned weights.

This weighting mechanism effectively balances the need for temporal relevance (by assigning the highest priority to recent data) with the retention of foundational historical information (by assigning a positive weight to all $N_R$ update intervals). 
This makes the approach particularly useful in nonstationary environments where underlying trends change gradually.

Algorithm~\ref{alg:recent_replay_decay} details the weighted sampling process.
\begin{algorithm}[htb]
\caption{Recent Replay with Decay Set Generation}
\label{alg:recent_replay_decay}
\begin{algorithmic}[1]
\State \textbf{Input:} Historical dataset $\mathcal{D}_H = \{(\mathbf{x}(t), \mathbf{y}(t)) \mid t_{-1} < t \le t_{n}\}$, Replay buffer size $S_R$, Number of recent updates $N_R$, Number of updates $n$
\State \textbf{Output:} Replay dataset $\mathcal{D}_R$\Procedure{GenerateRecentReplayDecaySet}{$\mathcal{D}_H$, $S_R$, $N_R$, $n$}\Comment{Executed after the $n$-th model training/update is completed. Let $t_{-1} \equiv 0$.}
\State $\mathcal{D}_R \gets \varnothing$ 
\Comment{Initialize the Replay dataset}
    \If{$n \leq N_R-1$} \Comment{Case 1: Replay all available tasks $m \in \{0, \dots, n\}$}
        \State $M_{\text{range}} \gets \{0, 1, \dots, n\}$
        \State $C \gets (n+1)(n+2) / 2$ \Comment{Normalization constant: $\sum_{i=1}^{n+1} i$}
    \Else \Comment{Case 2: Replay recent $N_R$ tasks $m \in \{n-N_R+1, \dots, n\}$}
        \State $M_{\text{range}} \gets \{n-N_R+1, \dots, n\}$
        \State $C \gets N_R(N_R+1) / 2$ \Comment{Normalization constant: $\sum_{i=1}^{N_R} i$}
\EndIf
\For{$m$ in $M_{\text{range}}$} 
\Comment{Iterate through tasks to be replayed}
\State $\mathcal{D}_{\text{Task}} \gets \{(\mathbf{x}(t), \mathbf{y}(t)) \in \mathcal{D}_H \mid t_{m-1} < t \le t_{m}\}$ \Comment{Data specific to task $m$}
    \If{$n \le N_R$}
        \State $W \gets m+1$ \Comment{Weight increases linearly from 1 to $n+1$}
    \Else
        \State $W \gets m - (n - N_R)$ \Comment{Weight increases linearly from 1 to $N_R$}
    \EndIf
    \State $S_{\text{Task}} \gets \text{Round}\left(W \cdot S_R / C\right)$ \Comment{Calculate required sample size for task $m$}
    \State $S_{\text{Sample}} \gets \min(S_{\text{Task}}, |\mathcal{D}_{\text{Task}}|)$ \Comment{Ensure we don't sample more than available}
    \If{$S_{\text{Sample}} > 0$}
        \State $\mathcal{D}_{\text{Sample}} \gets \text{UniformRandomSample}(\mathcal{D}_{\text{Task}}, S_{\text{Sample}})$
        \State $\mathcal{D}_R \gets \mathcal{D}_R \cup \mathcal{D}_{\text{Sample}}$
    \EndIf
\EndFor
\State \textbf{return} $\mathcal{D}_R$
\EndProcedure
\end{algorithmic}
\end{algorithm}
\end{document}